\documentclass{article}

\usepackage[sglblindworkshop, final]{latinx_2025}

\usepackage[utf8]{inputenc} 
\usepackage[T1]{fontenc}    
\usepackage{hyperref}       
\usepackage{url}            
\usepackage{booktabs}       
\usepackage{amsfonts}       
\usepackage{nicefrac}       
\usepackage{xcolor}         
\usepackage{subfigure}
\usepackage{graphicx}
\usepackage{amsmath}

\title{Measuring Similarity between Artistic and AI Generated Images using Siamese Neural Networks}

\author{%
  Diego Castro Elvira, Navil Pineda Rugerio, Jesús García-Ramírez\\  \textbf{Cecilia Reyes-Peña, Ricardo Ramos-Aguilar}\\
  UPIIT, Instituto Politécnico Nacional \\
  Guillermo Valle 11, Centro, C.P. 90000 Tlaxcala de Xicohténcatl, Tlaxcala, Mexico.\\
  \texttt{ \{dcastroe2100,npinedar2100\}}@alumno.ipn.mx \{jegarciara,creyespe,rramosa\}@ipn.mx
}

\begin{document}

\maketitle

\begin{abstract}
AI-generated art has sparked debates around potential plagiarism, as these images may closely resemble existing artworks. This research quantifies the similarity between original pieces and AI-generated counterparts, particularly those produced by the Stable Diffusion XL Refiner 1.0. We use Siamese Networks with frozen CLIP encoders and cosine similarity optimized through triplet loss. A dataset of paired original and generated images was built using image-to-image generation and custom prompts, enriched with semantic descriptors and BLIP-2 captions. Prior studies report up to 81\% style replication and 90\% visual similarity. Our results show high discriminative performance: training accuracy reached 99.9\%, and the best model configuration achieved 99.4\% test accuracy with strong inter-class separation ($\delta \mu$ = 0.677), demonstrating the effectiveness of our semantic-visual embeddings.
\end{abstract}

\section{Introduction}

The use of Artificial Intelligence (AI) in artistic creation has grown significantly, raising concerns about the originality of AI-generated works and their potential resemblance to human-made art—particularly in digital media, where models such as Stable Diffusion~\cite{rombach2022high} can produce highly realistic and stylized images. This study arises in the interest of improve the authentication processes of digital artworks, facilitating the detection of similarities between original images and those generated by AI. For this purpose, we have chosen an approach based on Siamese Convolutional Neural Networks, capable of comparing images based on semantic descriptors and measuring their similarity.

Recent advances in generative artificial intelligence, particularly in models such as Stable Diffusion, have raised concerns about copyright infringement due to their ability to replicate distinctive artistic styles. Studies show that these models can mimic the work of professional artists with up to 81\% accuracy in style imitation and 90\% visual similarity to original pieces~\cite{messuring}. To address these challenges, tools such as ArtSavant~\cite{moayeri2024artcopyright} and StyleAuditor~\cite{styleauditor} have been developed, integrating approaches based on CLIP, DeepMatch, and pretrained convolutional networks to detect stylistic imitation with accuracies above 89\%. Other works highlight the legal risks of unintended copyright violations, reporting that nearly 70\% of AI-generated content from generic prompts shows significant similarity with copyrighted materials~\cite{oncopyright}.

Beyond these studies, research on Siamese Neural Networks (SNNs) demonstrates their effectiveness in visual comparison tasks relevant to copyright detection. SNN-based models have achieved up to 94\% mean Average Precision in applications such as image retrieval, pattern recognition, and person re-identification across cameras~\cite{hybridcnn, tobacoo800, reid}. Their ability to generalize to unseen datasets and maintain robustness under noisy labeling makes them a strong candidate for evaluating stylistic similarity and potential infringements in AI-generated artworks.

In this context, to apply our proposed method we build different datasets using artistic images from WikiArt~\cite{wikiart} and their AI-generated counterparts produced with Stable Diffusion XL Refiner 1.0. 
\footnote{The datasets are available at \url{https://huggingface.co/Dant33}}.
The first dataset that we create its named as Wikiart\_with\_StableDiffusion, includes 81,444 AI-generated images derived from original paintings using prompts created with BLIP2~\cite{li2023blip} and LLaMA 3 8B~\cite{grattafiori2024llama}, and transformed using Stable Diffusion XL Refiner 1.0~\cite{esser2024scaling}.

The second dataset, WikiArt-81K-BLIP\_2-captions, consists of the original 81,444 WikiArt images enriched with genre labels, cleaned metadata, and image descriptions generated with BLIP2. The third dataset, WikiArt-81K-BLIP\_2-1024x1024, contains the same set of images resized uniformly to 1024×1024 pixels using LANCZOS resampling. The fourth dataset, WikiArt-81K-BLIP\_2-768x768, includes the same images resized to 768×768 pixels with padding to preserve aspect ratio. These datasets were developed to support research in AI-generated art analysis, image similarity, and style transfer.

A Siamese Neural Network with frozen CLIP encoders and triplet-based contrastive learning was proposed to analyze the semantic and visual similarity between original and generated artworks. The best model configuration achieved a triplet accuracy of 99.4\%, with a marked separation between the similarity scores of matched and mismatched image pairs. Through this work, we seek to provide a tool for the assessment of originality in AI-generated art, supporting artists and the research community in the identification of possible similarities.



Given the scarcity of specialized models for generating and evaluating similarity across paintings of different artistic genres, this work introduces a new approach that combines image-to-image generation with similarity assessment using Siamese Convolutional Neural Networks.

To support this study, we compiled a dataset containing both original artworks and their AI-generated counterparts across multiple styles. Unlike previous studies~\cite{elgammal2017can}, this research covers a broader and more diverse range of artistic genres. The associated code is publicly available in a GitHub repository~\footnote{\href{https://github.com/DiegoCastr00/CalculoSimilitud}{https://github.com/DiegoCastr00/CalculoSimilitud}}.

\section{Measuring the similarity between Artistic and AI-Generated images}

We follow a methodology structured in four stages, to implement and evaluate a similarity comparison model between original artworks and their AI-generated counterparts.

In its initial stages, this study is exploratory, since it seeks to establish relationships between the visual features of the original and generated images. Each stage of the process, from data preparation to model evaluation, is detailed in the following subsections of this document.

\subsection{Acquisition and cleaning of the WikiArt dataset}

In this stage of the project a systematic process is carried out for the preparation of data based on artistic works from WikiArt. Let denote $ I_{o} =  \big\{ I_{o}^{1}, I_{o}^{2}, \dots,I_{o}^{N} \big\}$ as the original set of images collected from Wikiart, where $I_{o}^{i}$ represents a digitized artistic work.

Each image $I_{o}^{i} \in I$ is associated with a metadata vector composed of: 
a genre, an artist and a phash value (an identifier containing the unique perceptual characteristics of each artwork to detect similarities and duplicates). The incomplete data that were not useful for the model was defined, cleaned or supplemented to obtain a representative data set. 

In order to adapt the data set to the architectural requirements of the Stable Diffusion XL Refiner 1.0 model, it was applied a transformation including resizing and aspect ratio adjustment. Since the model operates efficiently with resolutions of $1024 \times 1024$ and $768 \times 768$ pixels, the Lanczos resampling algorithm was used as the interpolation method. To maintain the original aspect ratio, conditional fill techniques were applied such as: zero padding, reflection, clamping and repetition.

Since the most common dimensions were in a range of width sizes between $1,381 px$ and $15,754 px$, and a height range between $1,381px$ and $17,768px$, 
the downsizing towards resolutions of $1024 \times 1024$ and $768 \times 768$ pixels did not imply a drastic decrease in the visual quality of the images, so it was feasible.


\subsection{Image generation with Stable Diffusion}

In the next stage, a series of images was generated using Stable Diffusion XL Refiner 1.0. For this purpose, customized prompts were constructed from the textual descriptions of the original works, using a BLIP2. The prompts were designed at two levels of transformation, examples of each one are the next:

\begin{itemize}
    \item \textbf{Moderate Transformation:} Where alterations to the image 
    are limited to subtle changes. This is possible through variations in the color scheme, lighting, and artistic technique used.
    \item \textbf{Radical Transformation:} Where more drastic alterations are applied, such as recontextualizing the work, historical transposition, cultural reinterpretation, or technological transformation.
\end{itemize}

This process prevented a single transformation strategy from being
applied to all images, resulting in a more diverse dataset. The Stable Diffusion XL Refiner 1.0 model was run using the generated textual descriptions. Each prompt was associated with a reference image, and the model generated the corresponding image that was stored along with its associated prompt. Examples of prompts include the next:

\begin{itemize}
    \item \textit{Radical Transformation.} \textbf{ \textit{``A futuristic, neon-lit cityscape where the cockatoo is now a cybernetic being, perched atop a glowing, holographic vase. The original's soft, dreamy atmosphere is replaced with a sense of high-tech urgency and innovation.''}}\\
    \item \textit{Moderate Transformation.} \textbf{\textit{``A whimsical Rococo scene, shifting the color palette to soft pastels and warm golden lighting, as if set during a moonlit summer evening. Reimagine the background as a lush, verdant garden, complete with twinkling fireflies''}\\}
\end{itemize}

\subsection{Triplet construction and Siamese Network Design}

For the training of the Siamese Neural Network model, a strategy based on comparison learning using image triplets was implemented. Each triplet is composed of three visual elements: an anchor image $I_a$, a positive image $I_p$, and a negative image $I_n$, formally defined as $T = (I_a, I_p, I_n)$. Where:

\begin{itemize}
    \item $I_a \in \mathbb{R}^{H \times W \times 3}$ is the anchor image, corresponding to an original artwork from WikiArt (with $H$ and $W$ being the image dimensions in pixels). \\
    \item $I_p \in \mathbb{R}^{H \times W \times 3}$ is the positive image, generated from $I_a$ using \textit{Stable Diffusion}. \\
    \item $I_n \in \mathbb{R}^{H \times W \times 3}$ is the negative image, randomly selected from a different class, style, or artistic context, with no direct visual or semantic relationship to $I_a$.
\end{itemize}

\begin{figure}[htbp]
\includegraphics[width=0.9\textwidth]{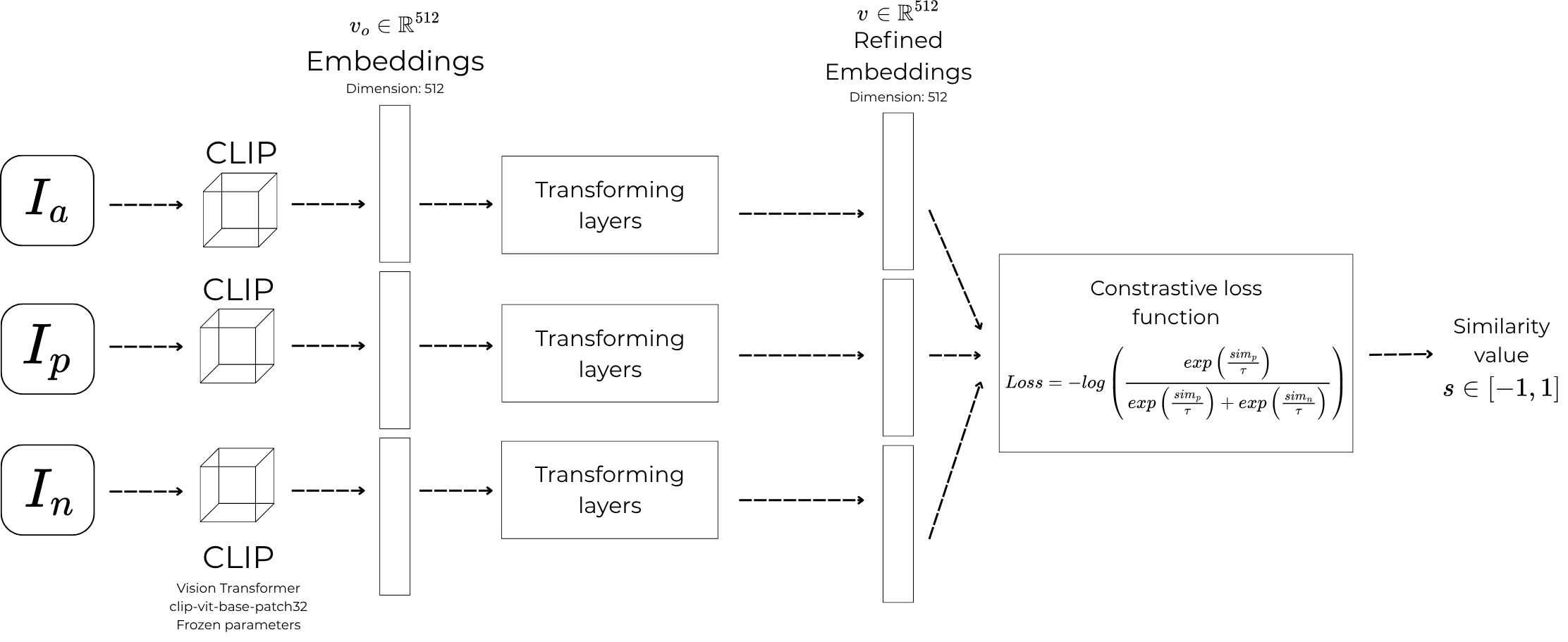}
\caption{Proposed Siamese Neural Network Architecture. Contrastive learning approach using image triplets.} \label{siamesearchitecture}
\end{figure}

The architecture designed for comparing similarity between artistic images is based on a Siamese Neural Network. A contrastive learning approach is employed using triplets $(I_a, I_p, I_n)$, as we can see in the Figure~\ref{siamesearchitecture}.

Each image in the triplet is processed through the pretrained CLIP model. This model divides the image into patches, processes them via multi-head attention and feed-forward layers, and finally generates a 512-dimensional vector that encapsulates general visual semantics.

This encoder remains frozen during training, leveraging its zero-shot capability learned from hundreds of millions of image–text pairs. Thus, for each image $I^i$, an initial representation $v_0 \in \mathbb{R}^{512}$ is obtained.

The output vectors from CLIP are not directly optimized for fine-grained comparison between generated and original images. Therefore, a trainable refinement block is introduced, consisting of:

\begin{itemize}
    \item A normalization layer that standardizes the input values.
    \item Adaptive transformers, implemented as transformer encoder layers with multi-head attention, residual connections, and non-linear feed-forward networks (with GELU activations).
    \item Linear projection and regularization, using trainable linear layers followed by dropout to prevent overfitting.
    \item Final L2 normalization $(\parallel v \parallel _2=1)$.
\end{itemize}

This transforms the vectors $v_0 \rightarrow v$. With the refined embeddings $v_a, v_p, v_n$, cosine similarity is computed between them as $sim_p = \cos(v_a, v_p), \quad sim_n = \cos(v_a, v_n)$.


The model is trained using a contrastive loss function based on InfoNCE~\cite{rusak2024infonce}, which encourages higher similarity between $v_a$ and $v_p$, and lower similarity between $v_a$ and $v_n$ as we can see in the equations~\ref{eq_loss}.

\begin{equation}
    \text{Loss} = -\log \left( \frac{\exp\left(\frac{sim_p}{\tau}\right)}{\exp\left(\frac{sim_p}{\tau}\right) + \exp\left(\frac{sim_n}{\tau}\right)} \right)
    \label{eq_loss}
\end{equation}

\subsection{Evaluation}

To quantify the model performance, an evaluation strategy based on triplet comparison was applied. The main metric used to compare embeddings was cosine similarity.
Each triplet receives two similarity scores:
\begin{itemize}
    \item $s^+ = s(v_a, v_p)$: similarity between the original image and its generated counterpart.
    \item $s^- = s(v_a, v_n)$: similarity between the original image and a negative sample.
\end{itemize}

A prediction is considered correct when the similarity between the anchor and the positive sample exceeds that of the anchor and the negative sample, i.e., when \( s^{+} > s^{-} \). The proportion of triplets satisfying this condition defines the \textit{Triplet Accuracy} metric, which serves as the primary indicator of model performance. This metric can be formally expressed using an indicator function \(\mathbf{1}(\cdot)\), defined to return 1 if the condition inside is true and 0 otherwise:

\begin{equation}
    \text{Triplet Accuracy} = \frac{1}{N} \sum_{i=1}^{N} \delta(s^+_i > s^-_i)
    \label{eq:triplet}
\end{equation}

where \(\mathbf{1}(s_i^{+} > s_i^{-})\) denotes the indicator function that evaluates to 1 when the similarity between the anchor and positive image is greater than that between the anchor and negative image, and 0 otherwise.

In addition to accuracy, the mean values of positive and negative similarities were computed, denoted as $E[s^+]$ and $E[s^-]$, respectively, as well as their mean separation $\delta s = E[s^+] - E[s^-]$. This value quantitatively indicates the model’s ability to discriminate between similar and dissimilar pairs.

\section{Experimental Results}
This section presents the outcomes of the data collection, curation, and dataset structuring process for training and evaluating the Siamese model for image similarity.

\subsection{Dataset description}

In our experiments we create a new dataset that contains 81,445 images were obtained, each corresponding to AI-generated images based on original artworks from various artistic genres. These genres include Impressionism, Minimalism, Cubism, Abstract Expressionism, among others. 

To support the experiments, we generated four datasets, each designed with specific purposes and metadata formats. Below is a summary of each dataset and its structure.

\begin{enumerate}
    \item \textbf{Wikiart with StableDiffusion}. This dataset is organized by grouping AI-generated images by artistic genre and includes metadata linking each generated image to its original WikiArt painting, along with the prompts used and their compressed versions.
    

    \item \textbf{WikiArt 81K BLIP-2 Captions}. This dataset is provided as a CSV file, where each row contains structured metadata and a description generated by BLIP-2 model for 81,000 images, including details such as artist, genre, painting title, perceptual hash, and subset (train or test).

    \item \textbf{WikiArt 81K BLIP-2 1024x1024}. This dataset replicate the same structure as the caption dataset but include images resized to 1024 pixels, allowing analysis at different resolutions. This version is designed for high-resolution analysis. It includes columns as —filename, genre, artist, painting name, perceptual hash, automatically generated description, and subset—but filenames have a “resize1024” suffix to distinguish them.

    \item \textbf{WikiArt 81K BLIP-2 768x768}. This dataset has the same structure as Dataset 3 but includes images resized to 768 pixels and with filenames carrying a “resize768” suffix to indicate the resolution.
\end{enumerate}

\subsection{Quantitative results}

\begin{figure}[htbp]
    \centering
    \subfigure[]{\includegraphics[width=0.48\linewidth]{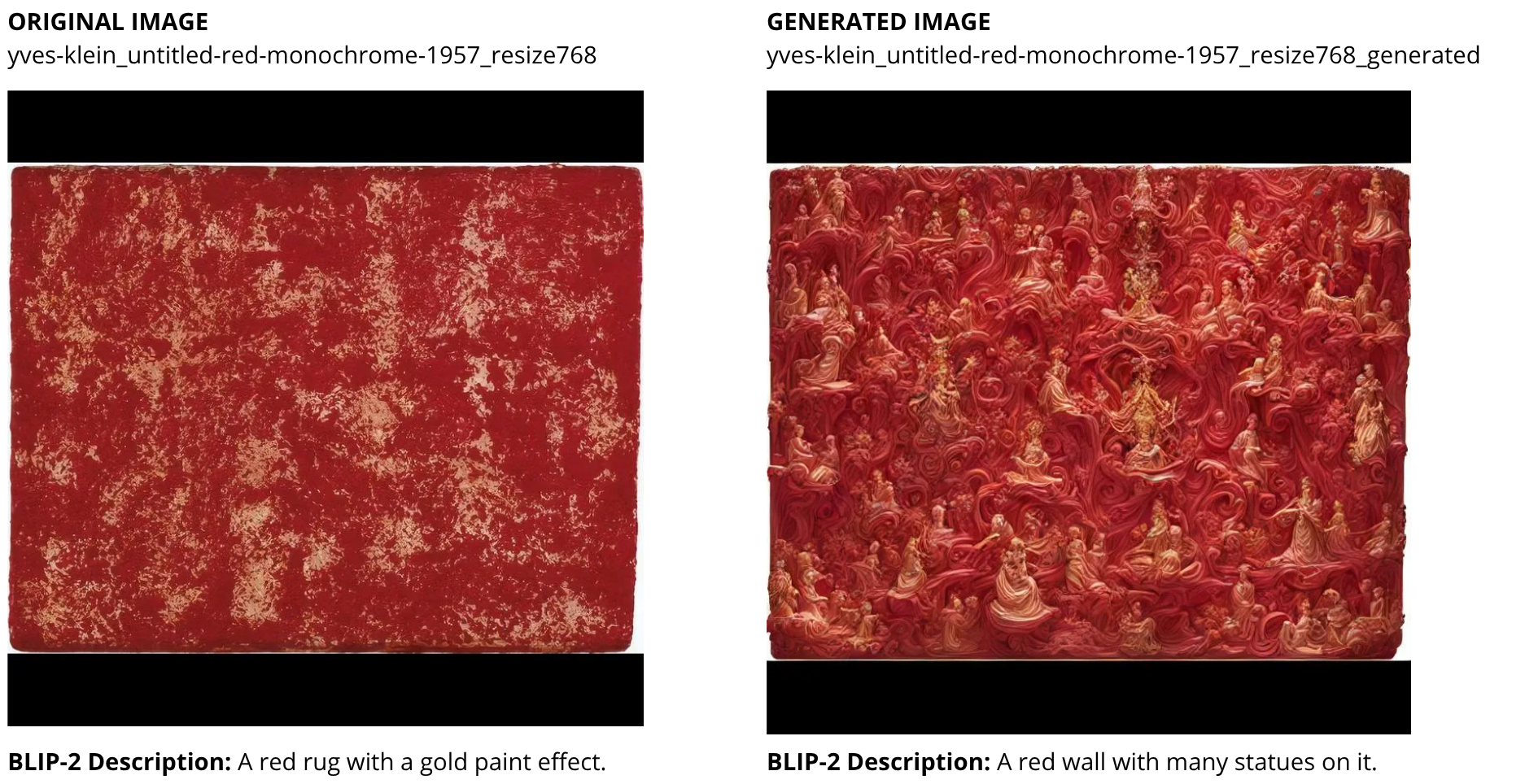}}
    \subfigure[]{\includegraphics[width=0.48\linewidth]{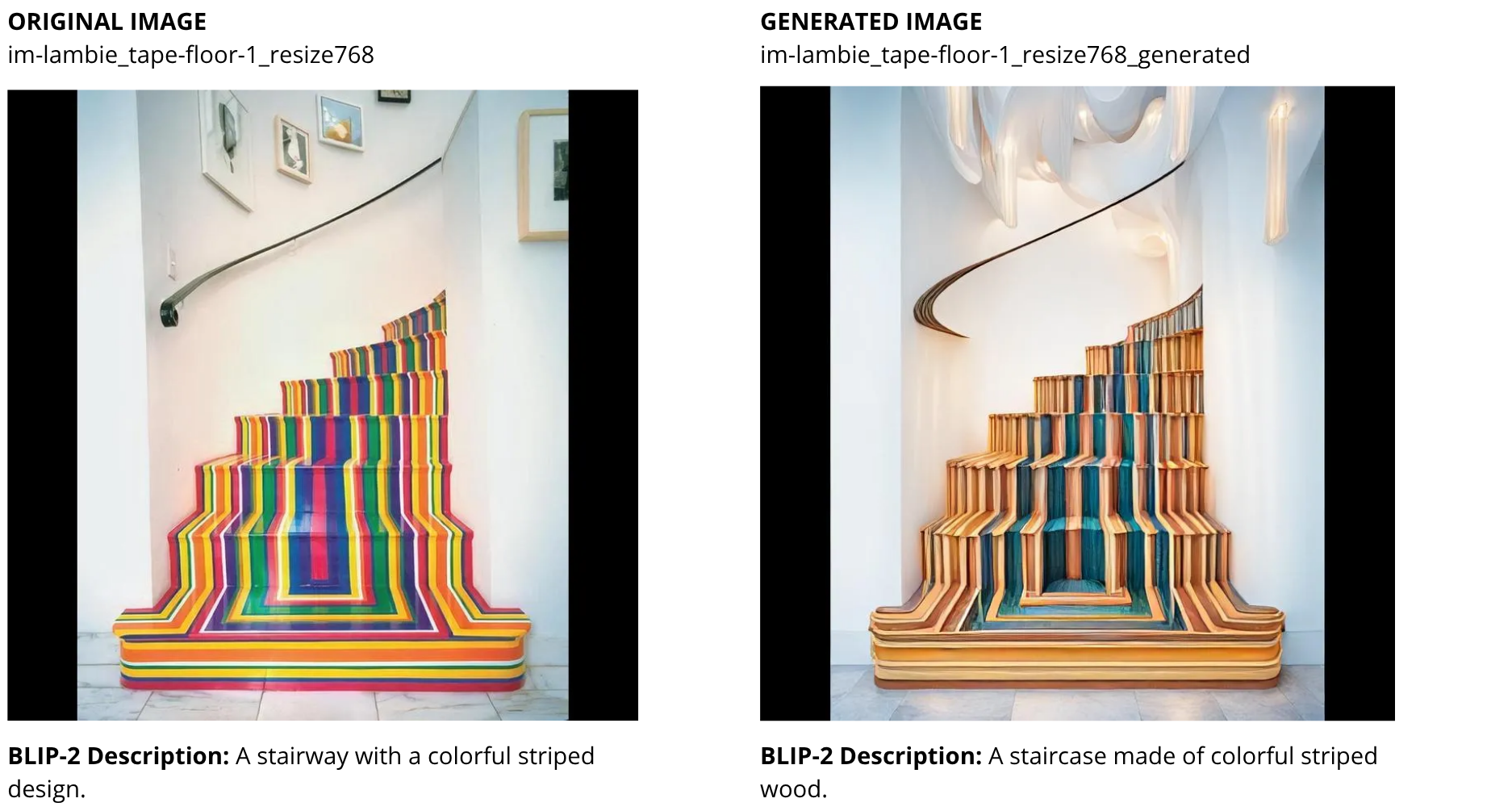}}
    \subfigure[]{\includegraphics[width=0.48\linewidth]{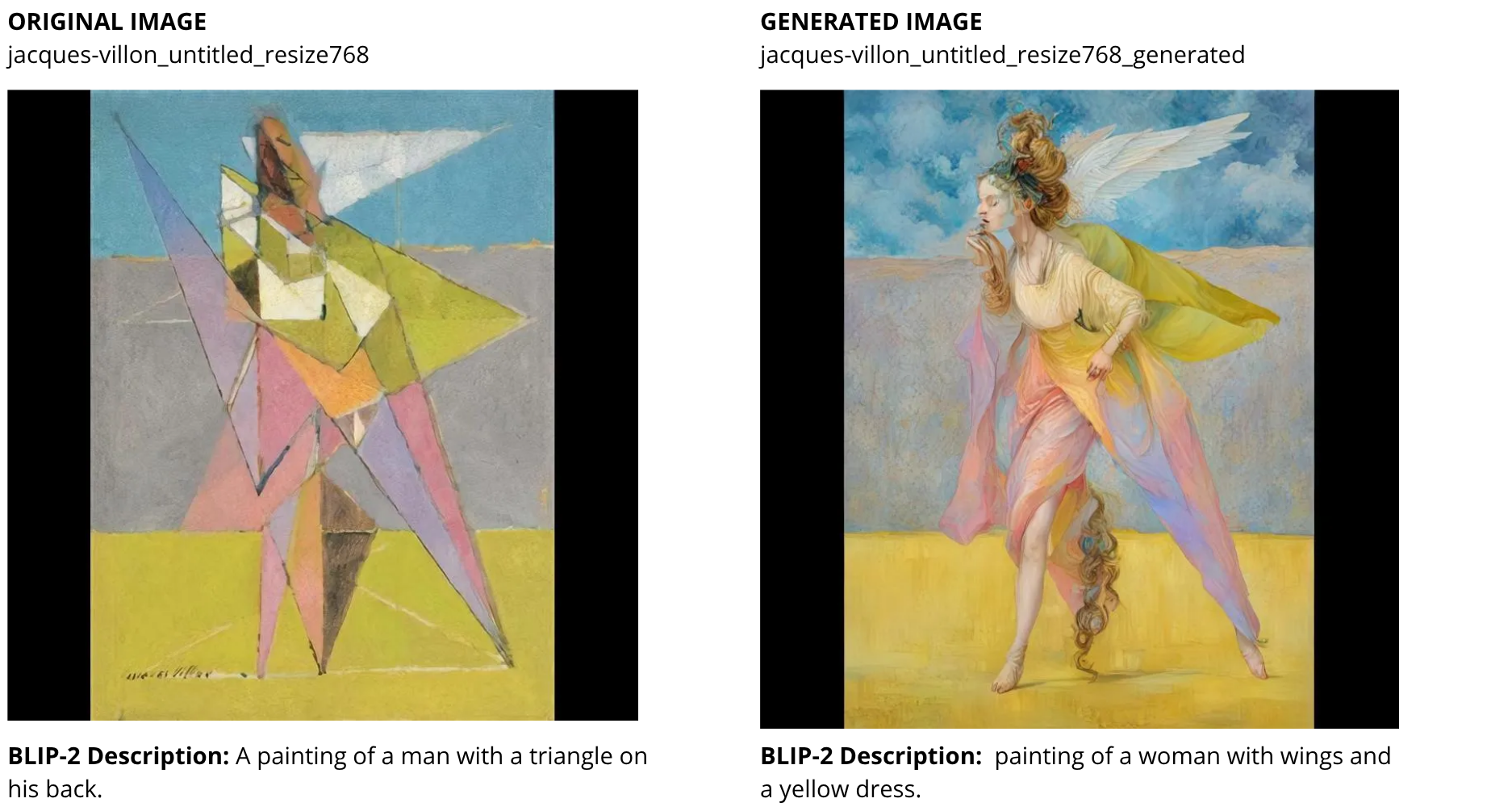}}
    \subfigure[]{\includegraphics[width=0.48\linewidth]{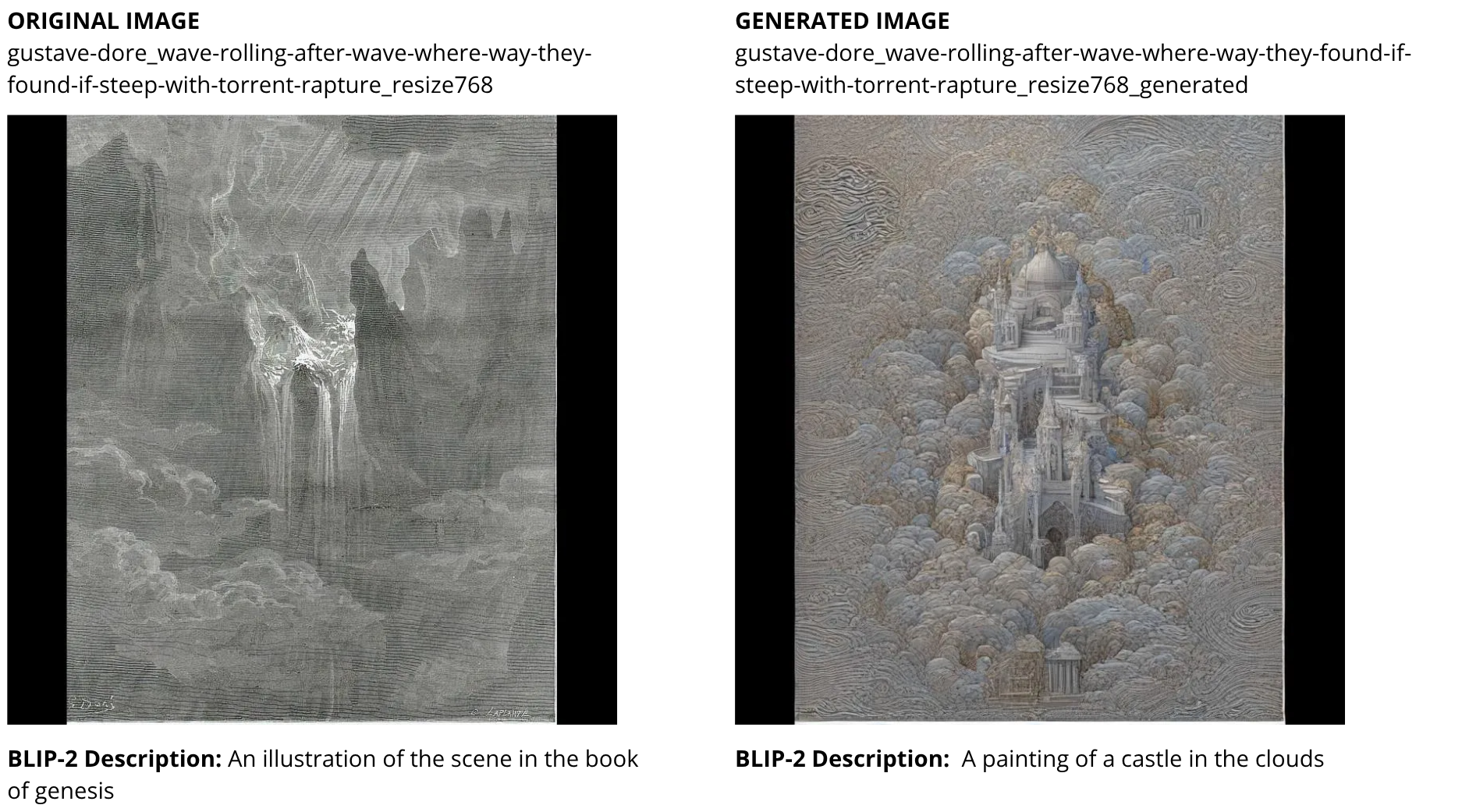}}
    \caption{Examples of generated images with the original images and the used prompt}
    \label{fig:examples}
\end{figure}

We present illustrative examples of artistic images used in our study. Each example includes the original artwork from WikiArt, accompanied by its title and a natural language description generated by the BLIP-2 model. We also show the corresponding image generated by Stable Diffusion XL Refiner 1.0 based on the original, along with the BLIP-2 description of the generated image. Examples of these images can be seen in Figure~\ref{fig:examples}.

These examples highlight how the semantic content is preserved or altered during the generation process. Comparing the descriptions provided by BLIP-2 for both original and generated images allows us to qualitatively assess the semantic alignment achieved by the model and to better understand the challenges in capturing artistic style and context in the generation process.

To generate images with Stable Diffusion, a reference image corresponding to the original artwork is required, along with a prompt that guides the diffusion model in creating the new image. These prompts were custom-designed and constructed based on the textual description of the original image, following two levels of transformation: moderate and radical.

\begin{figure}[htbp]
    \centering
    \subfigure[Original Image]{\includegraphics[width=0.30\linewidth]{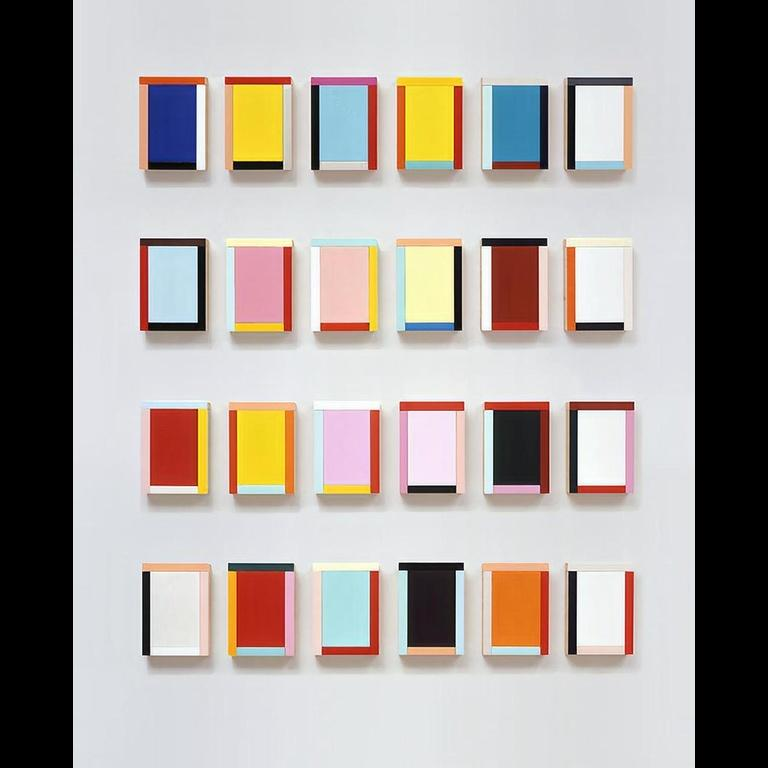}}
    \subfigure[Generated Image]{\includegraphics[width=0.30\linewidth]{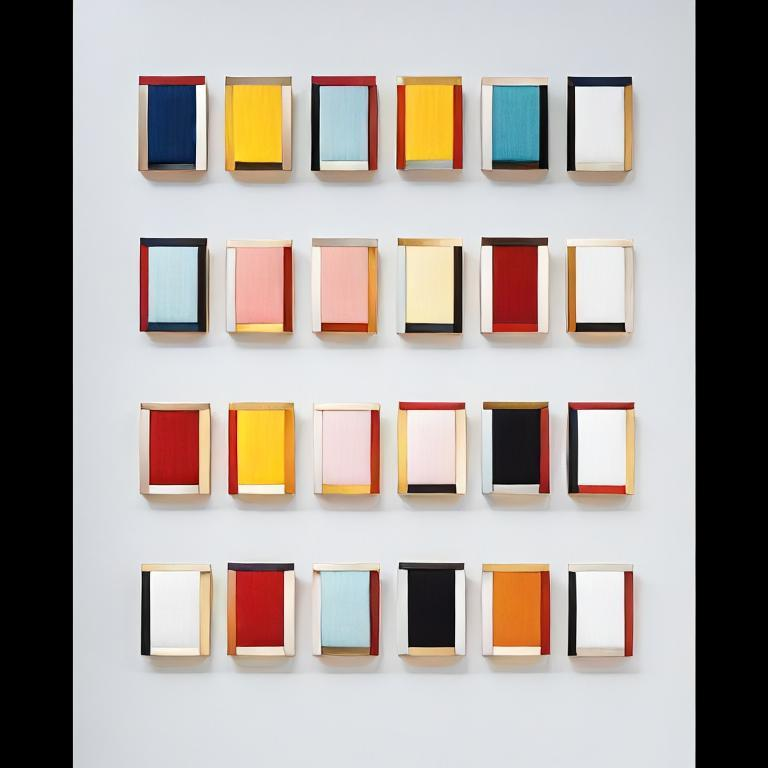}}
    \caption{Example of generated image from minimalist painting, this transformation shows only color changes.}
    \label{fig:generated_examples}
\end{figure}

At the moderate level, the transformation preserves the fundamental essence of the original image, introducing subtle yet noticeable changes that maintain the composition and main subject of the artwork. Modifications may include alterations in the color scheme, adjustments in lighting conditions, reinterpretations of artistic techniques, incorporation or modification of secondary elements, and style adaptations without losing the image core identity. An example of this configuration can be seen in the Figure~\ref{fig:generated_examples}.


On the other hand, the radical level involves a disruptive transformation of the image, recontextualizing it in such a way that only the original concept of the artwork is preserved. This approach encompasses ten distinct transformation strategies: historical transportation, cultural reinterpretation, natural or organic reimagination, abstract or surrealist deconstruction, emotional or psychological reframing, mythological or fantastical reinvention, technological reimagination, and scale modifications.


This prompt generation process was intentionally designed to avoid selecting a single transformation strategy per image, thereby diversifying the dataset. It also aims to ensure variety across transformations, providing a broad spectrum of creative reinterpretations. The key difference between the two transformation levels lies in the magnitude of change: while the moderate approach introduces variations within the same conceptual universe, the radical approach completely challenges the original perception, creating entirely new images conditioned on the original artwork.


To evaluate the generalization capability of the model, two different data split configurations were used during training:

\begin{itemize}
    \item \textbf{Configuration A:} 70\% training, 15\% validation, 15\% test
    \item \textbf{Configuration B:} 80\% training, 10\% validation, 10\% test
\end{itemize}

Each configuration enabled the comparison of the model performance under different levels of training data availability. All training procedures were executed using the same model architecture and training settings, previously described in the methodology section.

The model was trained using a mini-batch size of thirty-two to ensure a balance between computational efficiency and gradient stability. The learning rate was configured to $3 \times 10^{-4}$. To prevent overfitting and encourage generalization, a weight decay coefficient of $1 \times 10^{-4}$ was applied. The temperature parameter in the contrastive loss function was set to $0.07$, which controls the sharpness of the similarity distribution. All experiments were conducted on NVIDIA RTX 3090 GPUs with multi-GPU support, facilitating parallelized training and enabling efficient processing of large-scale triplet datasets.


\subsection{Training Results}

\begin{figure}[htbp]
    \centering
    \subfigure[Train loss]{\includegraphics[width=0.48\linewidth]{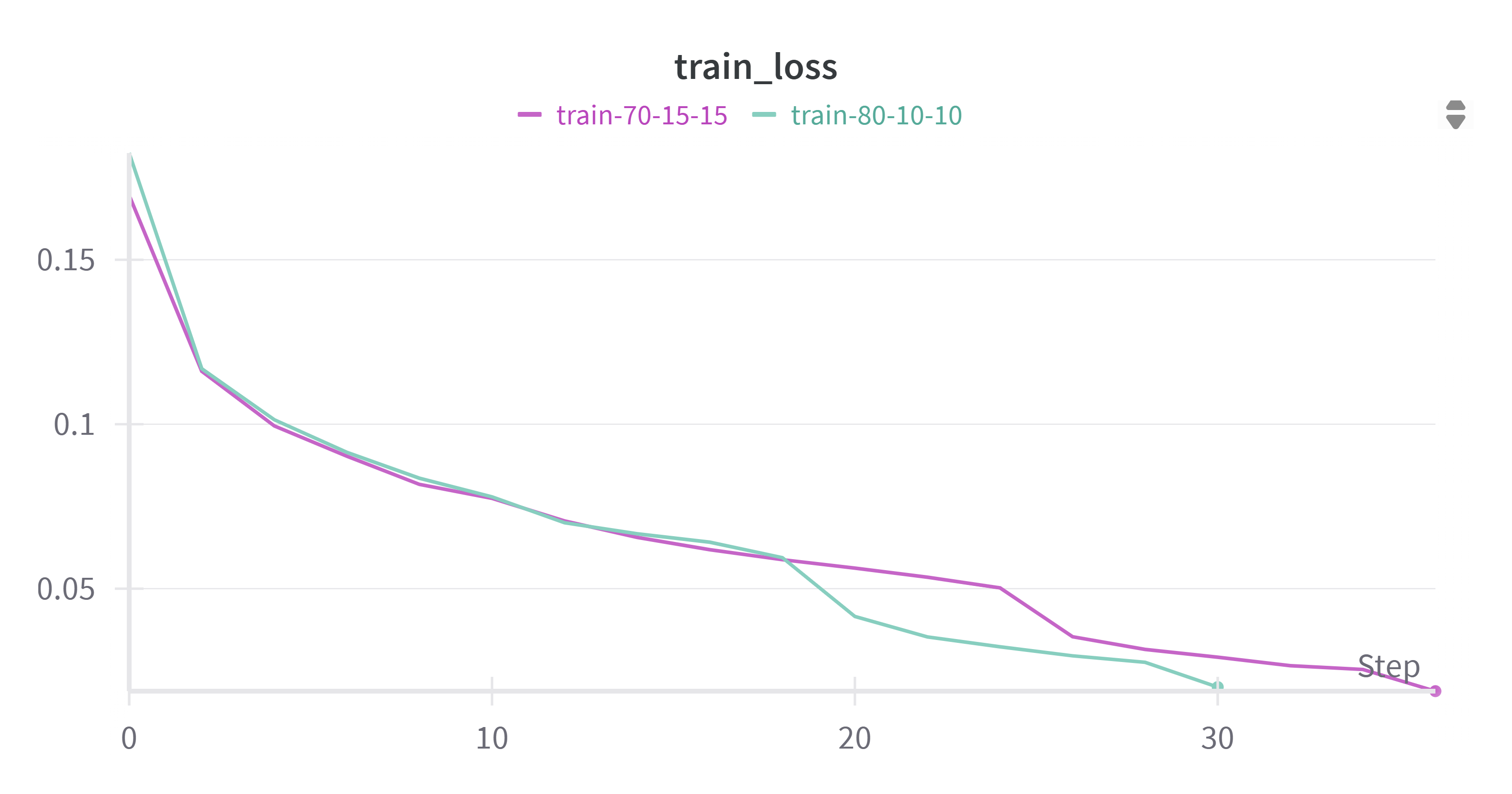}}
    \subfigure[Validation loss]{\includegraphics[width=0.48\linewidth]{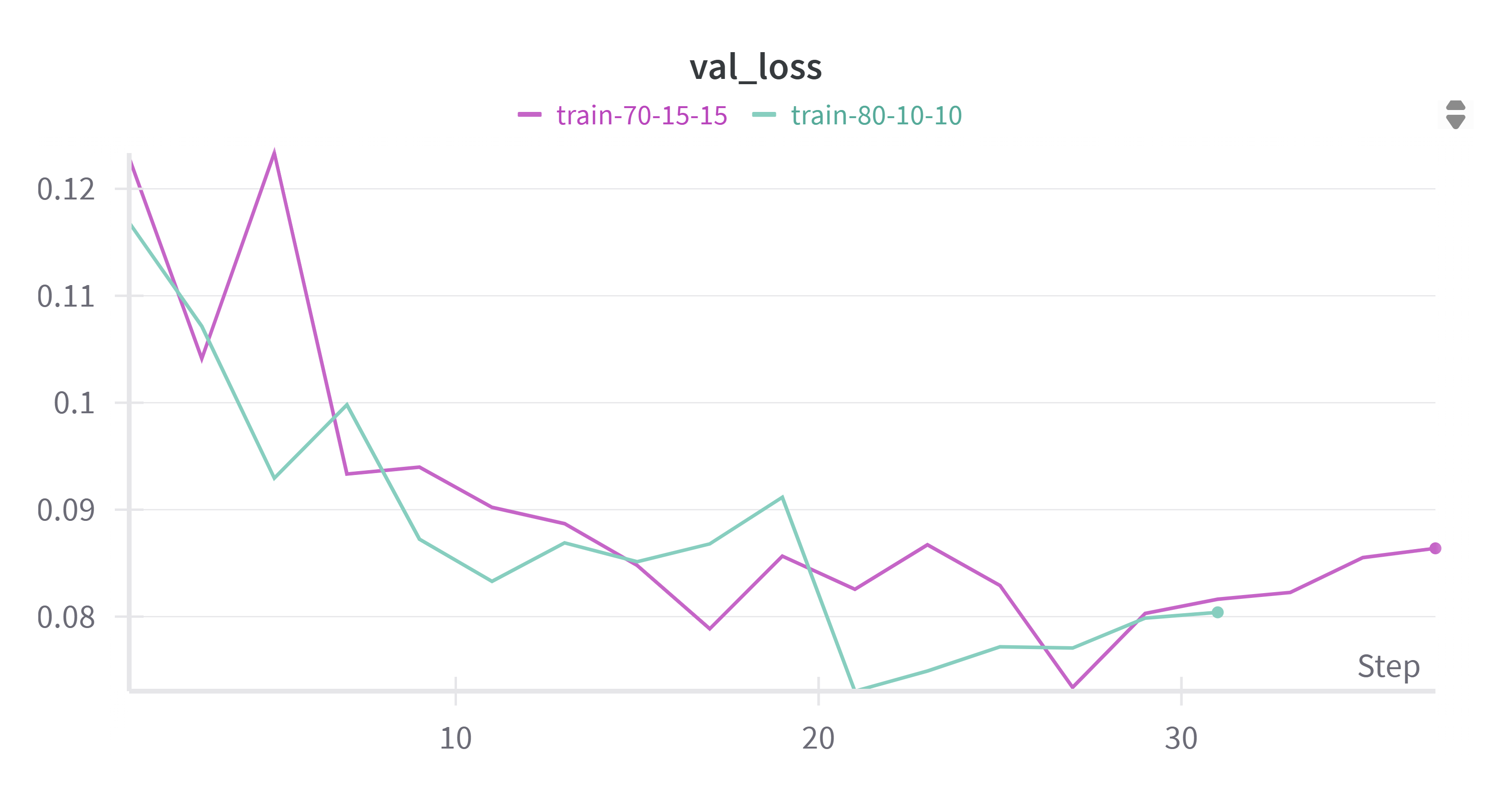}}
    \caption{Train and validation loss function for both model configurations.}
    \label{fig:train_val_loss}
\end{figure}

Figure~\ref{fig:train_val_loss} (a) shows that the evolution of the loss function on the training set starts around ($\approx 0.17$) and decreases rapidly during the first five epochs, followed by a smoother decline, converging towards a value close to $0.02$.

In contrast, the validation loss, as shown in Figure~\ref{fig:train_val_loss} (b), exhibits a more oscillatory behavior. Although a downward trend is present, several peaks appear, particularly in the early epochs, which likely reflect the model’s adaptation to the data.

\begin{figure}[htbp]
    \centering 
    \subfigure[Train triplet accuracy]{\includegraphics[width=0.48\linewidth]{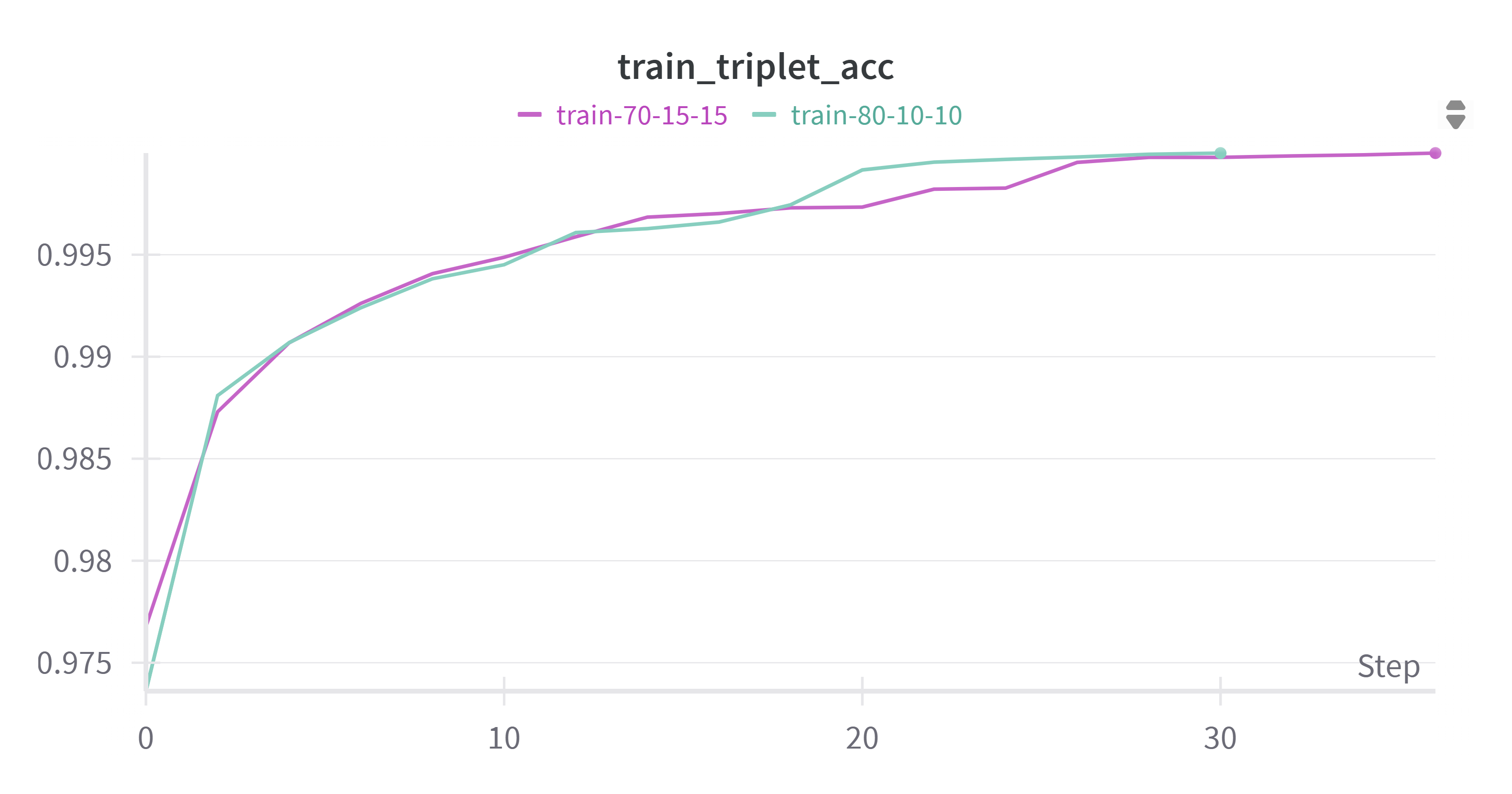}}
    \subfigure[Validation triplet accuracy]{\includegraphics[width=0.48\linewidth]{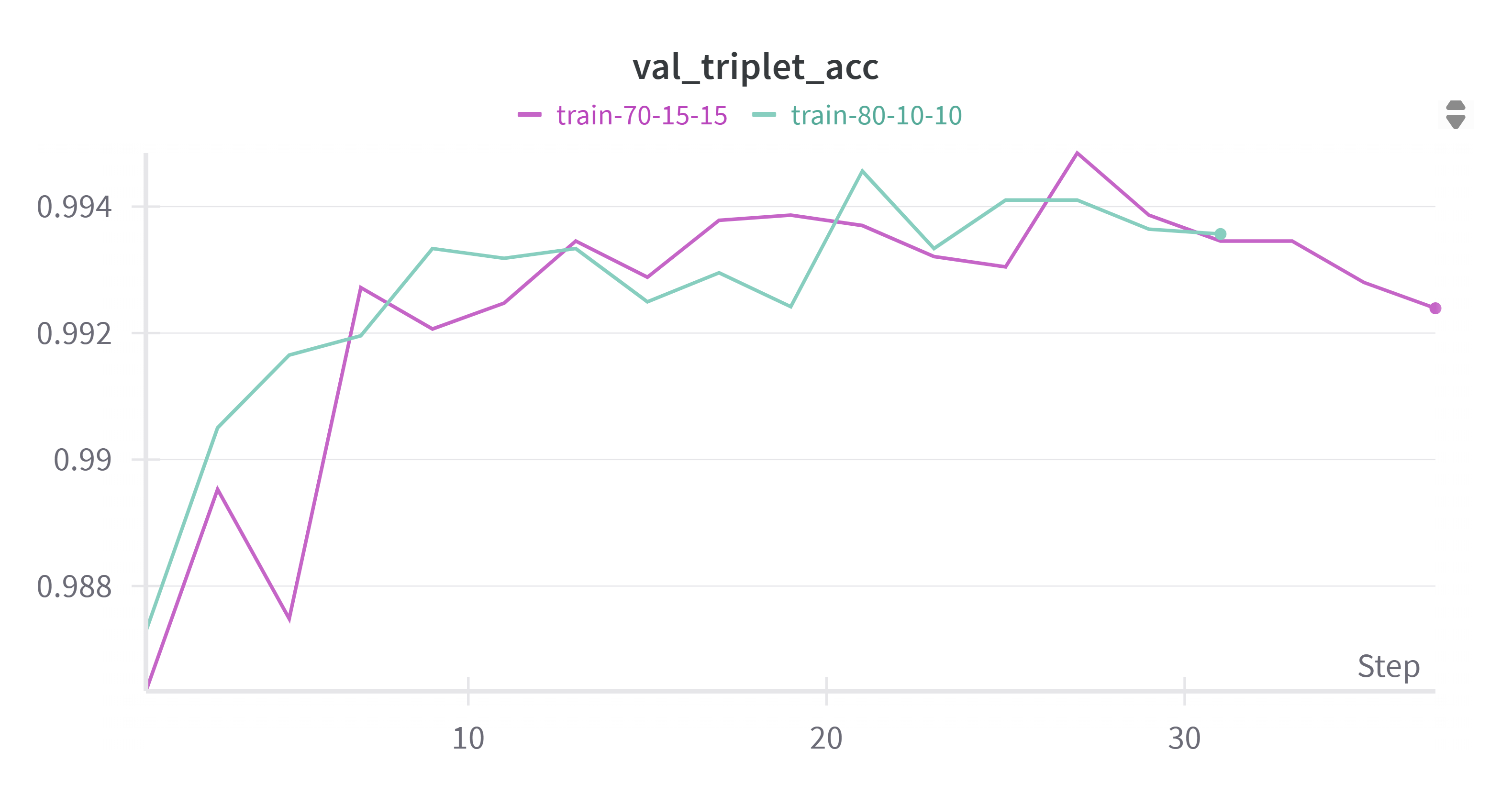}}
    \caption{Train and validation triplet accuracy for both model configurations.}
    \label{fig:train_val_triplet_acc}
\end{figure}

Regarding triplet accuracy, the train triplet accuracy metric starts at a high value ($\approx 97.5\%$), and as seen in Figure~\ref{fig:train_val_triplet_acc} (a), improves rapidly during the initial epochs, reaching values above 99\% and stabilizing around 99.9\%.


The validation triplet accuracy metric behaves similarly to the training metric. As shown in Figure~\ref{fig:train_val_triplet_acc} (b), it starts around 98.7\% and increases to a maximum of approximately 99.5\%. However, a slight decline is observed in the final stages.

These observations suggest that the model has a strong learning and convergence capacity. No signs of severe numerical instability were identified. However, analyzing these metrics together indicates that the optimal training point likely lies between epochs 20 and 25, before overfitting starts degrading generalization performance. Therefore, this interval is considered optimal for final model selection.



Both model configurations were evaluated, and the quantitative results are summarized in the Table~\ref{tab:evaluation}. The results show that both configurations achieve high performance, with accuracies above 98\%. However, the model trained with the 70-15-15 split achieved the best results in both triplet accuracy and average inter-class separation $\Delta \mu$. This greater separation between distributions indicates that the learned representations are more discriminative.

\begin{table}[htbp]
\centering
\caption{Evaluation of the two Siamese model configurations.}
\label{tab:evaluation}
\begin{tabular}{lcc}
\hline\hline
Configuration & Triplet Accuracy & $\boldsymbol{\Delta \mu = \mu_\text{pos} - \mu_\text{neg}}$ \\
\hline
\textbf{70-15-15} & \textbf{0.994} & \textbf{0.677} \\
80-10-10 & 0.985 & 0.463 \\
\hline
\end{tabular}
\end{table}

\begin{figure}[htbp]
    \centering
    \includegraphics[width=0.5\linewidth]{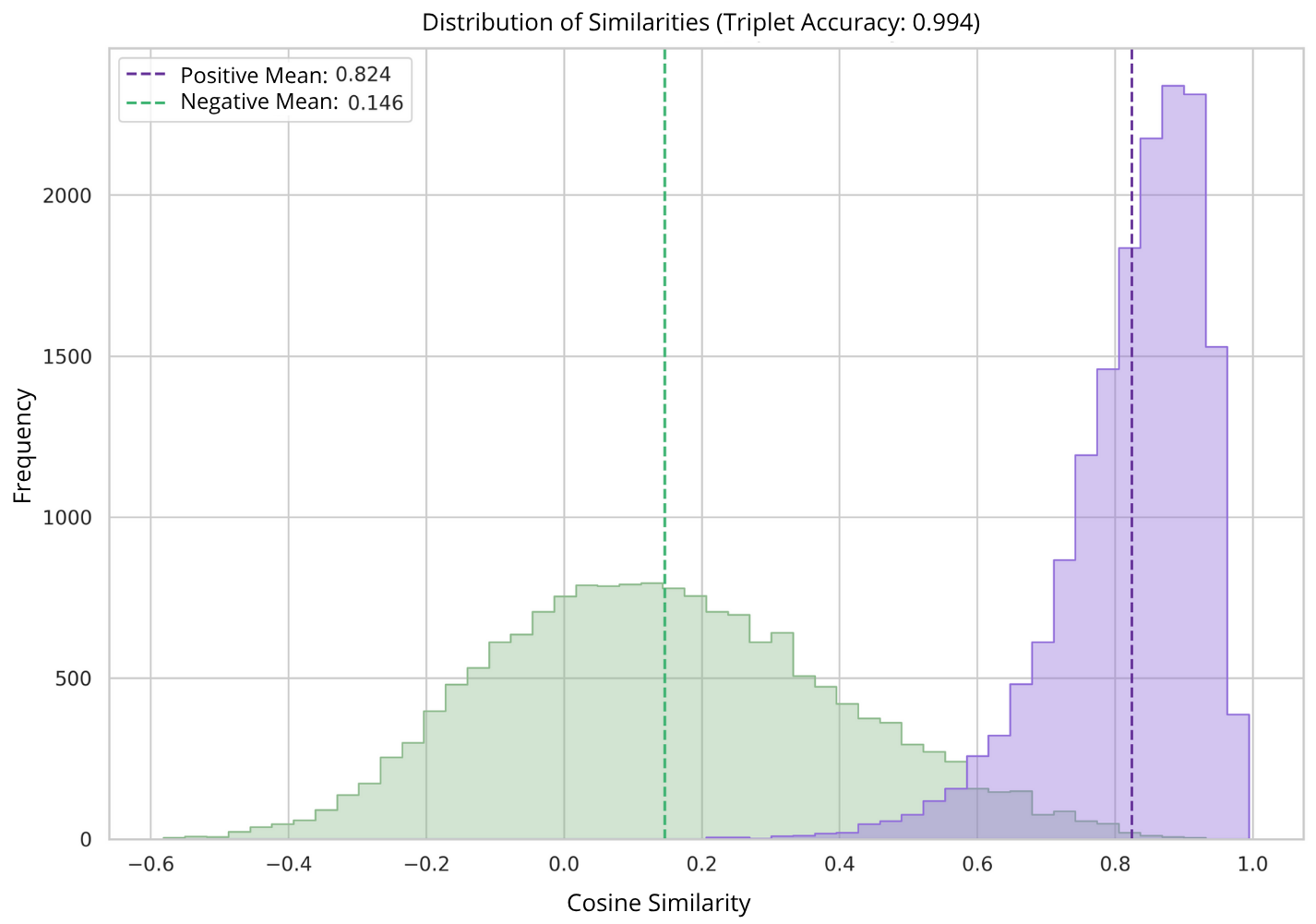}
    \caption{Similarity histogram of the best-trained model.}
    \label{fig:test-histogram}
\end{figure}

Figure~\ref{fig:test-histogram} shows exactly the expected behavior: positive similarities cluster near 1, while negative ones tend towards lower values, closer to 0. The minimal overlap between both curves indicates that a clear distinction between the two types of similarities is achieved.

To complement the training and validation metrics, an inference-based evaluation was performed to qualitatively analyze the model behavior when comparing pairs of images. This method allows us to observe how the model responds to visual similarity at test time, beyond aggregate performance metrics. Three representative use cases are presented to illustrate the model capacity to capture semantic similarity.

\begin{figure}[htbp]
    \centering
    \subfigure[Case 1]{\includegraphics[width=0.47\linewidth]{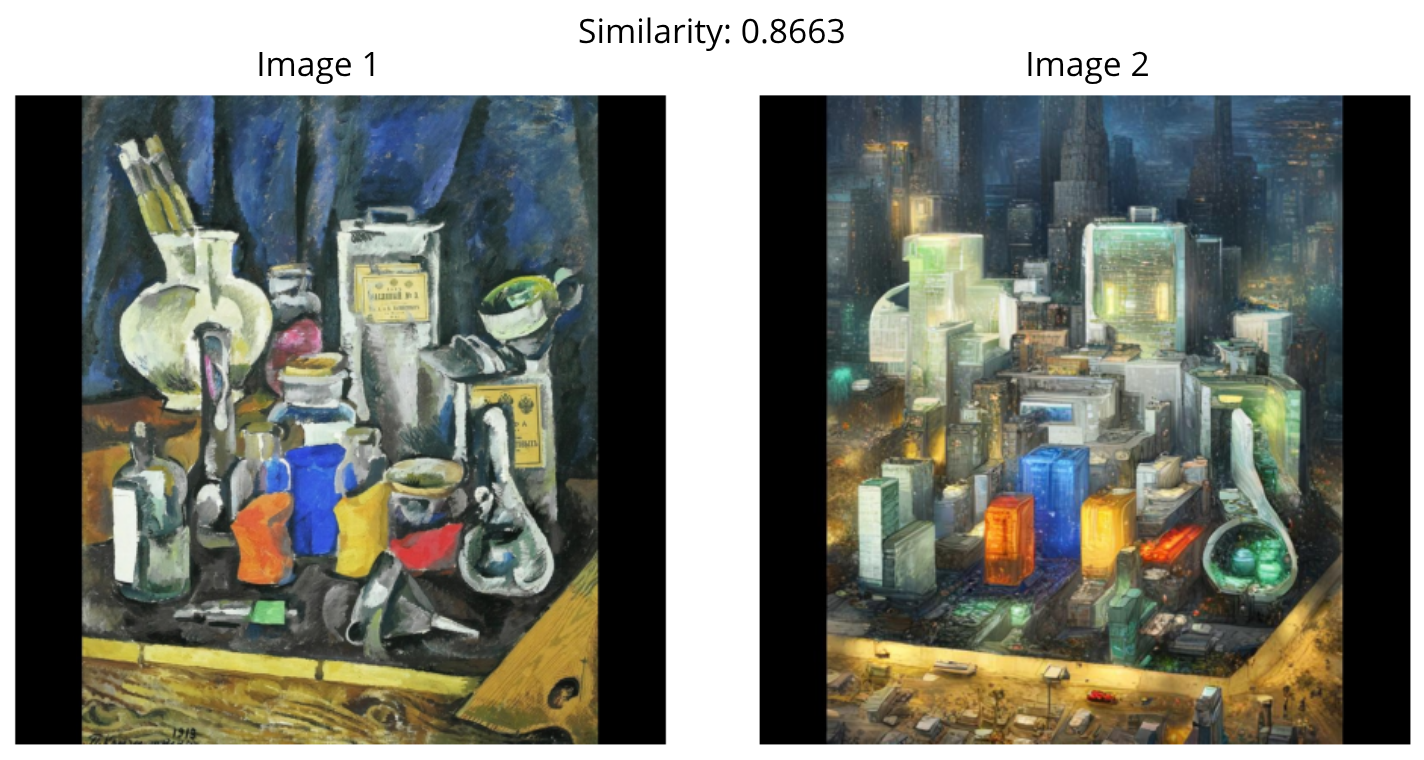}}
    \subfigure[Case 2]{\includegraphics[width=0.47\linewidth]{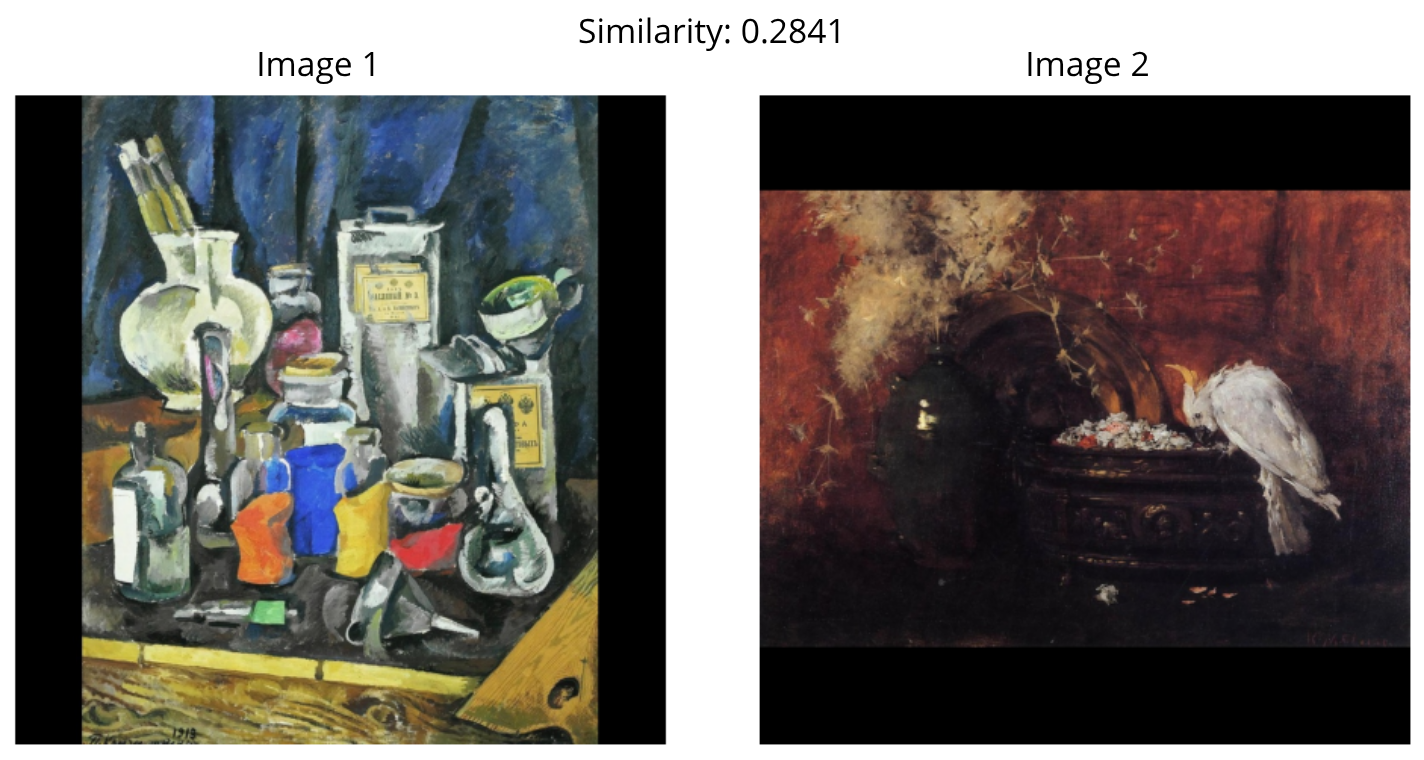}}
    \subfigure[Case 3]{\includegraphics[width=0.47\linewidth]{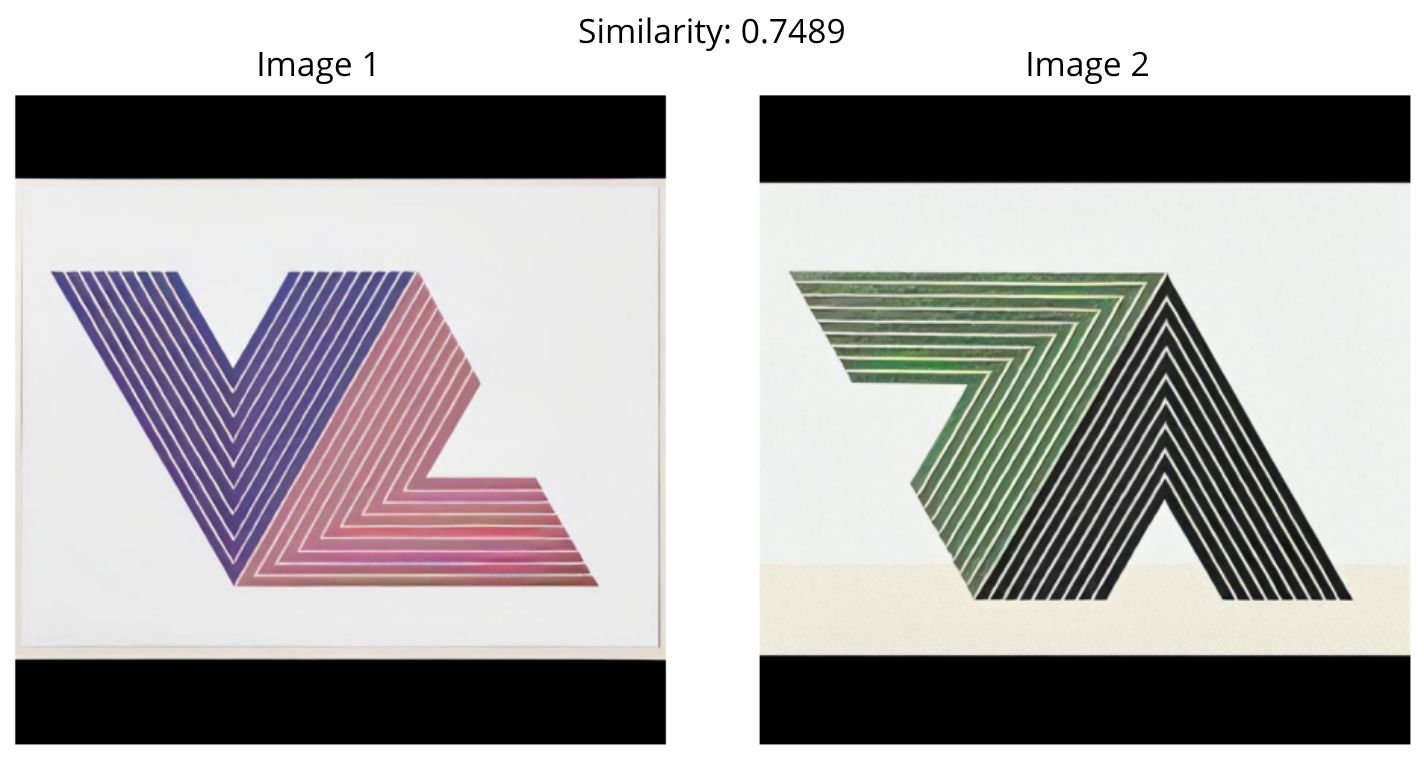}}

    \caption{Inference Case 1,2, and 3.}
    \label{fig:cases}
\end{figure}

In the Case 1 we compare the original and the generated counterpart images. A pair consisting of an original artwork and its counterpart generated using Stable Diffusion. As expected, the similarity score computed by the model is close to 1, indicating that the learned embedding space successfully preserves semantic alignment between conceptually equivalent images. This is illustrated in Figure~\ref{fig:cases} (a). 


The Case 2 compare unrelated images. A pair composed of completely unrelated images. The similarity score in this case is close to 0 or even negative, reflecting the model ability to discriminate between semantically dissimilar contents. In Figure~\ref{fig:cases} (b) it is shown the similarity value is $0.2841$.
    

Finally, in the Case 3 we compare similar images with style variations. This pair includes two visually similar images, where one has been rotated and recolored. Although they share core structural and semantic features, these superficial transformations challenge the model. The resulting similarity score is moderate—neither high nor close to zero—demonstrating the model's partial robustness to certain visual distortions. See Figure~\ref{fig:cases} (c), as shown the similarity value is $0.7489$.


These examples provide a qualitative insight into the discriminative power and limitations of the learned embedding space, especially in terms of invariance to transformations such as rotation and color manipulation.


\section{Conclusions and Future Work}

This work addresses the challenge of measuring semantic similarity between original artworks and their AI-generated counterparts. We trained a Siamese network in a semantic embedding space using contrastive triplet loss, with a CLIP-based backbone enhanced by learnable projection layers. The training process incorporated semantic descriptors from BLIP-2 and synthetic triplets generated via Stable Diffusion XL Refiner 1.0.

Our approach achieved a triplet accuracy of 99.34\% and demonstrated clear separation between positive and negative similarity distributions, indicating the learned embeddings capture meaningful semantic patterns. Results show that similarity learning, even without explicit class supervision, can effectively distinguish artistic styles.

We observed slight overfitting toward the end of training, suggesting that early stopping could improve generalization. The best-performing model confirms that stylistic similarity can be captured with minimal architectural adjustments to the Siamese framework.

Overall, this study highlights the potential of multimodal Transformer-based encoders for training similarity models that can help assess the fidelity and authenticity of AI-generated images. As generative models become more powerful and widespread, such tools may play a key role in detecting unauthorized stylistic imitation and verifying digital provenance in creative industries.

\section*{Acknowledgements}

The authors thankfully acknowledge computer resources, technical advice and support  provided  by  Laboratorio  Nacional  de  Supercómputo  del  Sureste  de México  (LNS),  a  member  of  the  SECIHTI  national  laboratories,  with  project No. 202501029N.

\bibliographystyle{plainnat}
\bibliography{bib}

@article{moayeri2024artcopyright,
  title={Rethinking Artistic Copyright Infringements in the Era of Text-to-Image Generative Models},
  author={Mazda Moayeri and Samyadeep Basu and Sriram Balasubramanian and Priyatham Kattakinda and Atoosa Chengini and Robert Brauneis and Soheil Feizi},
  journal={University of Maryland, Computer Science Department},
  year={2024}
}

@inproceedings{esser2024scaling,
  title={Scaling rectified flow transformers for high-resolution image synthesis},
  author={Esser, Patrick and Kulal, Sumith and Blattmann, Andreas and Entezari, Rahim and M{\"u}ller, Jonas and Saini, Harry and Levi, Yam and Lorenz, Dominik and Sauer, Axel and Boesel, Frederic and others},
  booktitle={Forty-first international conference on machine learning},
  year={2024}
}

@misc{ wikiart,
    author = "{WikiArt}",
    title = "WikiArt",
    howpublished = "\url{https://www.wikiart.org/}",
    note = "[Online; accessed 13-September-2024]"
  }

@article{styleauditor,
  title={WIP: Auditing Artist Style Pirate in Text-to-image Generation Models.},
  author={Du, L. and Zhu, Z. and Chen, M. and Ji, S. and Cheng, P. and Chen, J. and Zhang, Z.},
  journal={CISPA Helmholtz Center for Information Security, Saarbrucken},
  year={2024}
}

@article{messuring,
  title={Measuring the Success of Diffusion Models at Imitating Human Artists.},
  author={Casper, S. and Guo, Z. and Mogulothu, S. and Marinov, Z. and Deshpande, C. and Yew, R. J. and Dai, Z. and Hadfield-Menell, D.},
  journal={International Conference on Machine Learning},
  year={2023}
}

@article{oncopyright,
  title={On Copyright Risks of Text-to-Image Diffusion Models.},
  author={Zhang, Y. and Teoh, T. T. and Lim, W. H. and Wang, H. and Kawaguchi, K.},
  journal={National University of Singapore},
  year={2024}
}

@article{hybridcnn,
  title={Siamese Network Features for Image Matching.},
  author={Melekhov, I. and Kannala, J. and Rahtu, E.},
  journal={IEEE},
  year={2017}
}

@article{reid,
  title={Deep Siamese Network with Multi-level Similarity Perception for Person Re-identification.},
  author={Shen, C. and Jin, Z. and Zhao, Y. and Fu, Z. and Jiang, R. and Chen, Y. and Hua, X.},
  journal={IEEE},
  year={2017}
}

@article{tobacoo800,
  title={Image Retrieval and Pattern Spotting using Siamese Neural Network.},
  author={Wiggers, K. L. and Britto Jr., A. S. and Heutte, L. and Koerich, A. L. and Oliveira, L. S.},
  journal={ArXiv},
  year={2019}
}

@article{grattafiori2024llama,
  title={The llama 3 herd of models},
  author={Grattafiori, Aaron and Dubey, Abhimanyu and Jauhri, Abhinav and Pandey, Abhinav and Kadian, Abhishek and Al-Dahle, Ahmad and Letman, Aiesha and Mathur, Akhil and Schelten, Alan and Vaughan, Alex and others},
  journal={arXiv preprint arXiv:2407.21783},
  year={2024}
}

@inproceedings{li2023blip,
  title={Blip-2: Bootstrapping language-image pre-training with frozen image encoders and large language models},
  author={Li, Junnan and Li, Dongxu and Savarese, Silvio and Hoi, Steven},
  booktitle={International conference on machine learning},
  pages={19730--19742},
  year={2023},
  organization={PMLR}
}

@article{rusak2024infonce,
  title={InfoNCE: Identifying the Gap Between Theory and Practice},
  author={Rusak, Evgenia and Reizinger, Patrik and Juhos, Attila and Bringmann, Oliver and Zimmermann, Roland S and Brendel, Wieland},
  journal={arXiv preprint arXiv:2407.00143},
  year={2024}
}

@article{rombach2022high,
  title={High-resolution image synthesis with latent diffusion models},
  author={Rombach, Robin and Blattmann, Andreas and Lorenz, Dominik and Esser, Patrick and Ommer, Björn},
  journal={Proceedings of the IEEE/CVF Conference on Computer Vision and Pattern Recognition},
  year={2022},
  pages={10684--10695}
}

@article{elgammal2017can,
  title={CAN: Creative Adversarial Networks, generating “art” by learning about styles and deviating from style norms},
  author={Elgammal, Ahmed and Liu, Bingchen and Elhoseiny, Mohamed and Mazzone, Marian},
  journal={arXiv preprint arXiv:1706.07068},
  year={2017}
}





\end{document}